\documentclass[pdflatex,sn-mathphys-num]{sn-jnl}

\usepackage{graphicx}%
\usepackage{multirow}%
\usepackage{amsmath,amssymb,amsfonts}%
\usepackage{amsthm}%
\usepackage{mathrsfs}%
\usepackage[title]{appendix}%
\usepackage{xcolor}%
\usepackage{textcomp}%
\usepackage{manyfoot}%
\usepackage{booktabs}%
\usepackage{algorithm}%
\usepackage{algorithmicx}%
\usepackage{algpseudocode}%
\usepackage{listings}%
\usepackage{caption}
\usepackage{comment}
\usepackage{todonotes}

\usepackage{float}
\usepackage{array}

\theoremstyle{thmstyleone}%
\theoremstyle{thmstyletwo}%

\theoremstyle{thmstylethree}%

\begin{document}

\title[]{Auditable Clinical Timeline Reconstruction with Provenance-Aware Evidence Graphs}

\author*[1]{\fnm{Judith Jeyafreeda} \sur{Andrew}}\email{judithjeyafreeda@gmail.com}

\affil[1]{\orgdiv{Public Health Unit, Medical Information Department}, \orgname{Bordeaux University Hospital (CHU Bordeaux)}, \city{Bordeaux}, \country{France}}

\abstract{
Longitudinal clinical evidence is spread across notes and can be revised as new results arrive. A system that reconstructs a patient timeline is auditable only if it keeps the mentions behind each answer, records how facts were revised, and declines to answer when the evidence is not in the text. This study builds and tests these three properties in a fully synthetic setting (1,000 patients, 3,353 notes, 2,220 clinical facts, 220 revision edges). Two provenance-aware Evidence Graph operators reduced the node-plus-edge count to 67\% and 63\% of the uncompressed graph (77\% and 78\% of its serialized size) while preserving every answer and mention link across 6,813 sampled query points, all of which were answerable by recency; a fixed-window baseline reached 58\% of the size but returned no value for 53.4\% of points and a wrong value for 1.2\%, none flagged. Evidence-availability abstention was tested with two kinds of control. On controls carrying an explicit marker of omission, a BioClinicalBERT gate and a zero-shot large language model (LLM) gate both responded mainly to the marker: appending an omission notice to an otherwise unchanged note cut answering from 100\% to 3.4\% for BERT and from 58.9\% to 0\% for the LLM. Marker-free controls removed the temporal evidence with no notice, paired with a minimal rewrite by the same editor that preserved it. BERT abstained on 0 of 81 such notes while its accuracy on them fell from 93.8\% to 59.3\%; the drop occurred in all three relation classes and was not explained by the edit's reordering of event mentions. The LLM showed partial, marker-independent sensitivity: coverage fell from 75.6\% to 27.7\% on notes from which neither a cue filter nor the same model acting as judge could recover the relation. Against 482 gold evidence spans regenerated after paraphrasing, the LLM's cited spans reached recall 0.850 and precision 0.864 on answered notes, whereas BERT's evidence head, which received no span supervision, did not localize evidence (token-overlap recall 0.049 ungated, 0.006 gated). A temporally versioned provenance graph stored abstentions as typed, queryable edges with their reasons. The clean relation task admits a no-learning shortcut (0.651 accuracy), and all results describe implementation behaviour on synthetic data, not clinical performance or utility.
}

\maketitle
\section{Introduction}\label{sec:intro}
 
Longitudinal clinical information is spread across documents written at different times and for different purposes. A later culture result, imaging study, specialist assessment or treatment response can revise an earlier working diagnosis or treatment plan. A system that reports only the current answer hides this history: a reviewer cannot tell which mentions support the answer, when they were written, what they replaced, or whether the system had enough text to answer in the first place. Temporal relation extraction from clinical narratives remains difficult and depends on purpose-built annotated corpora~\cite{Andrew2026annotation}.
 
Auditability is treated here as three concrete properties of a timeline-reconstruction pipeline. First, the compressed representation keeps revision history and mention links, so that an answer to ``which fact was active on day~$d$'' can be traced to the notes it came from. Second, the extractor declines to answer when the text does not contain the evidence, instead of returning a relation drawn from priors or surface regularities. Third, declines are stored in the same versioned graph as answers, with their reason, so that a withheld answer can be told apart from a missing value. The pipeline builds on provenance tracking from database theory~\cite{Green2007provenance} and on selective classification~\cite{ElYaniv2010,Geifman2017}.
 
This is a synthetic mechanism-validation study, not a clinical performance evaluation. A synthetic corpus gives what the three properties need and real corpora do not provide: a known revision structure, known injection sites for contradictions, and the ability to remove evidence from a note under control. An earlier version of the pipeline gave strong compression results but an evidence-span recall of 0.000 in every extraction configuration. That value traced back to an evaluation defect: gold spans had been recorded before the notes were paraphrased and were searched for by exact match, so the metric could not say whether the gates were finding the right evidence. Two independent checks were therefore added. Gold spans were regenerated against the paraphrased text, and the gates were scored on inputs from which the relation-bearing evidence had been removed, where the correct behaviour is to abstain. The way those evidence-unavailable inputs are built turned out to decide how far the results can be read. The first control set announces the removal with a fixed sentence, an appended notice or a mask token, so a gate can pass it by reading the announcement. A second, marker-free set removes the evidence silently and is paired with a minimal rewrite that keeps it (Sections~\ref{sec:controls} and~\ref{sec:silent}).
 
The contributions are:
\begin{itemize}
\item Provenance-aware Evidence Graph operators that reduce representation size while preserving as-of queries and mention retention, evaluated against a fixed-window baseline and a generator-derived oracle, with size reported both as a node-plus-edge count and as serialized bytes.
\item An evidence-availability evaluation of two gate families, BioClinicalBERT and a zero-shot LLM, on marker-carrying and marker-free evidence-unavailable inputs, with metrics that separate clean coverage, abstention recall and answers given on unavailable evidence, and with each control paired with its clean note.
\item A temporally versioned provenance graph in which abstention is a first-class edge type, demonstrated on gated and ungated outputs of one model.
\item A documented list of failure modes: the invalid span-recall metric, gate over-abstention, an unattainable certification threshold, a template shortcut in the clean relation task, and explicit omission markers in the control inputs.
\end{itemize}
 
The study addresses three questions.
\begin{enumerate}
\item Do provenance-aware graph operators preserve answers and mention retention under compression?
\item Do evidence gates abstain when the evidence has been removed, both when the removal is announced and when it is not, and what does this cost on clean inputs?
\item Can a temporally versioned provenance graph store abstentions as first-class, queryable edges alongside answers?
\end{enumerate}
No claim is made about performance on real clinical records.

\section{Methods}

\subsection{Architecture overview}
\begin{figure}[t]
\centering
\includegraphics[width=\textwidth]{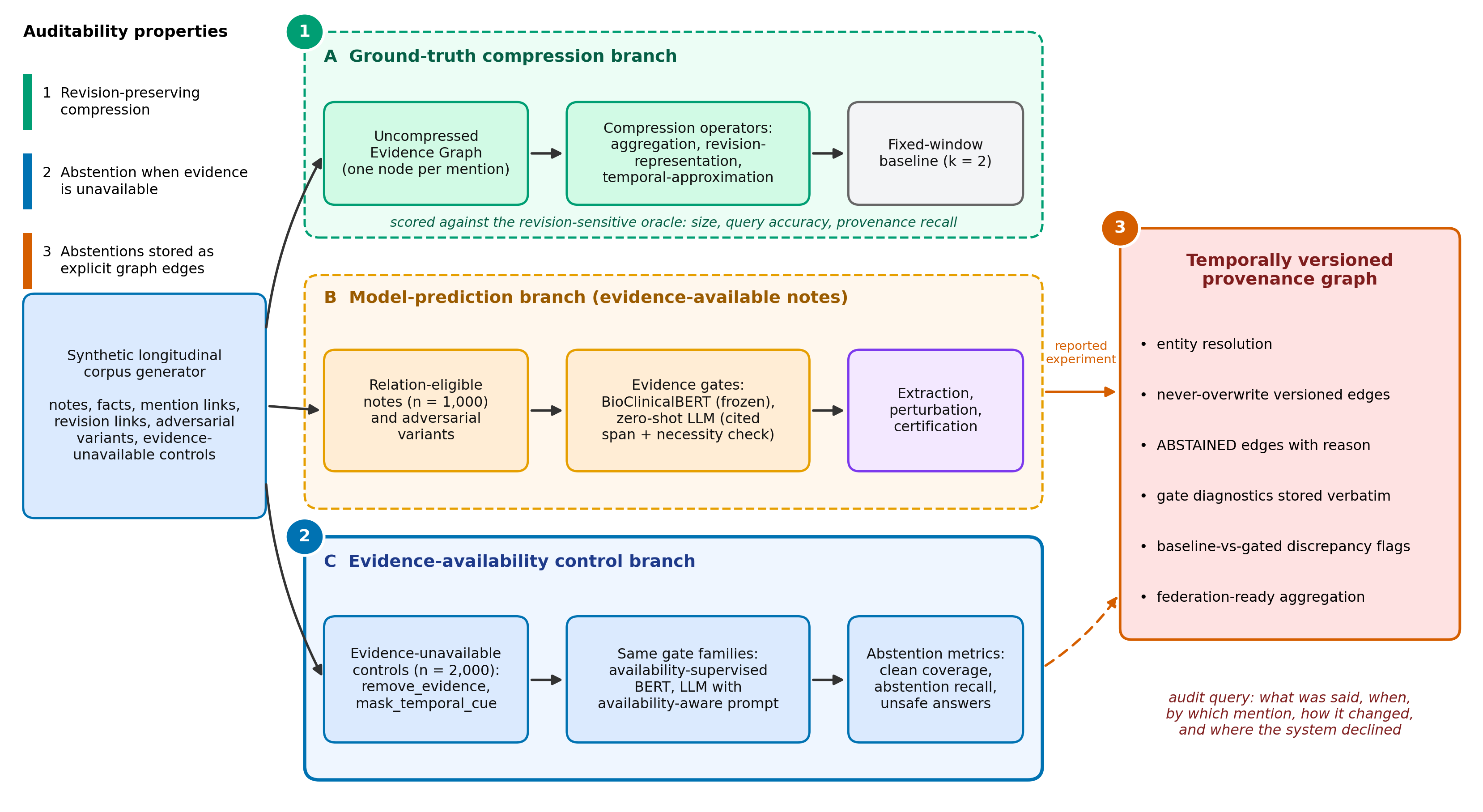}
\caption{Architecture of the synthetic mechanism-validation pipeline. The generator produces notes, facts, mention links, revision links, adversarial variants and evidence-unavailable controls. Branch A builds an Evidence Graph directly from generator facts and scores compression against a revision-sensitive oracle; extractor predictions do not enter it. Branch B runs the evidence gates on evidence-available notes. Branch C runs the gate families on evidence-unavailable controls. The provenance graph is independent of the compression graph. The solid arrow marks the reported experiment (ungated versus gated LLM outputs); the dashed arrow marks ingestion of control-branch outputs, which the infrastructure supports but which was not evaluated here.}
\label{fig:arch}
\end{figure}

Figure~\ref{fig:arch} shows three analytically separate branches. In the ground-truth compression branch (A), generator-produced facts, mention links, revision links and simulated confidence values are assembled into an uncompressed Evidence Graph and the compression operators are scored against generator-produced oracle answers. No extractor prediction enters this branch. In the model-prediction branch (B), relation-eligible notes are processed by a BioClinicalBERT pipeline~\cite{alsentzer2019} and by a zero-shot LLM pipeline (\texttt{mistralai/Mistral-Small-3.2-24B-Instruct-2506})~\cite{mistral2025}. The outputs are used for relation-extraction evaluation, perturbation analysis, certification and provenance-graph ingestion. In the evidence-availability control branch (C), the same gate families are run on inputs whose evidence the generator has removed or masked, and the outcome is scored by abstention.

Provenance recall is the fraction of original mention-node identifiers retained as targets of membership edges in the compressed representation. This measures mention retention through the implemented membership links and does not establish arbitrary graph reachability.

\subsection{Synthetic longitudinal corpus}\label{sec:corpus}
Patient timelines were generated programmatically with an initial working diagnosis and empiric treatment, followed in a subset of patients by a diagnostic or treatment revision. Each timeline requested 3--6 notes; because note days were drawn with replacement and coinciding days merged, the realized count was 1--6 notes per patient (median 3; 177 patients had fewer than three). Notes were rendered in four styles (terse ICU, verbose discharge, nursing, structured admission) by deterministic lexical transformation or by an LLM paraphrase request. The generator records fact identifiers, mention links, event days and revision links. Only notes with a non-null gold relation and gold evidence span entered relation classification, which gives one note per patient (1,000 notes; 354 \texttt{EVENT1-OVERLAP-EVENT2}, 345 \texttt{EVENT1-BEFORE-EVENT2}, 301 \texttt{EVENT2-BEFORE-EVENT1}, so a majority-class rate of 0.354). Additional revision and follow-up notes were kept for corpus-level and graph-level analyses.

The paraphrase prompt instructed the model to preserve every clinical fact, date and temporal relation, but semantic preservation was not independently validated. Successful API return is therefore recorded as paraphrase completion, not as proof of fact preservation. The LLM was used to generate synthetic note variants and not to create, alter or interpret the reported scientific conclusions.

An adversarial set was generated by injecting one or more typed contradiction clauses (fabricated symptom, temporal contradiction, dosage contradiction, negation flip) into selected base notes; the generator stores the injected spans and contradiction types. The generator also provides a revision-sensitive oracle that returns the active fact for each concept at a query day. The oracle is computed from fact days and revision pointers and is used only to evaluate the compression branch.

A synthetic corpus was chosen over public temporal-relation corpora (i2b2/n2c2 \cite{sun2013i2b2}, THYME \cite{styler2014thyme}, MIMIC-III/IV \cite{johnson2016mimic3, johnson2023mimic4}) for three reasons. The compression experiments need a programmatically known revision structure (which fact superseded which and when), which pairwise TLINK annotations do not provide. The manipulation and certification experiments need typed contradictions with known injection sites, which would require new manual annotation on a real corpus. And credentialed corpora carry data use agreements that restrict routing note text through a third-party paraphrasing API. Evaluation on a real, expert-annotated, DUA-compliant corpus is required future work before any clinical-performance claim.

Each relation-eligible note derives from one of three fixed templates, which differ in surface form as well as temporal content: the EVENT2-BEFORE-EVENT1 template mentions the treatment first, and the overlap template carries one day number rather than two. On the released notes, a no-learning rule using only mention order and the number of day mentions reaches 0.651 accuracy (0.633 on the BERT test split; 0.996 on structured-admission notes). In 336 of 646 BEFORE-type relations both events fall on the same day, and in the EVENT1-BEFORE-EVENT2 template the ordering cue attaches to the symptom report, not to the diagnosis. Clean accuracies should be read against this shortcut baseline as well as the majority-class rate (0.354).

\subsection{Evidence-unavailable controls}\label{sec:controls}
\begin{figure}[t]
\centering
\includegraphics[width=\textwidth]{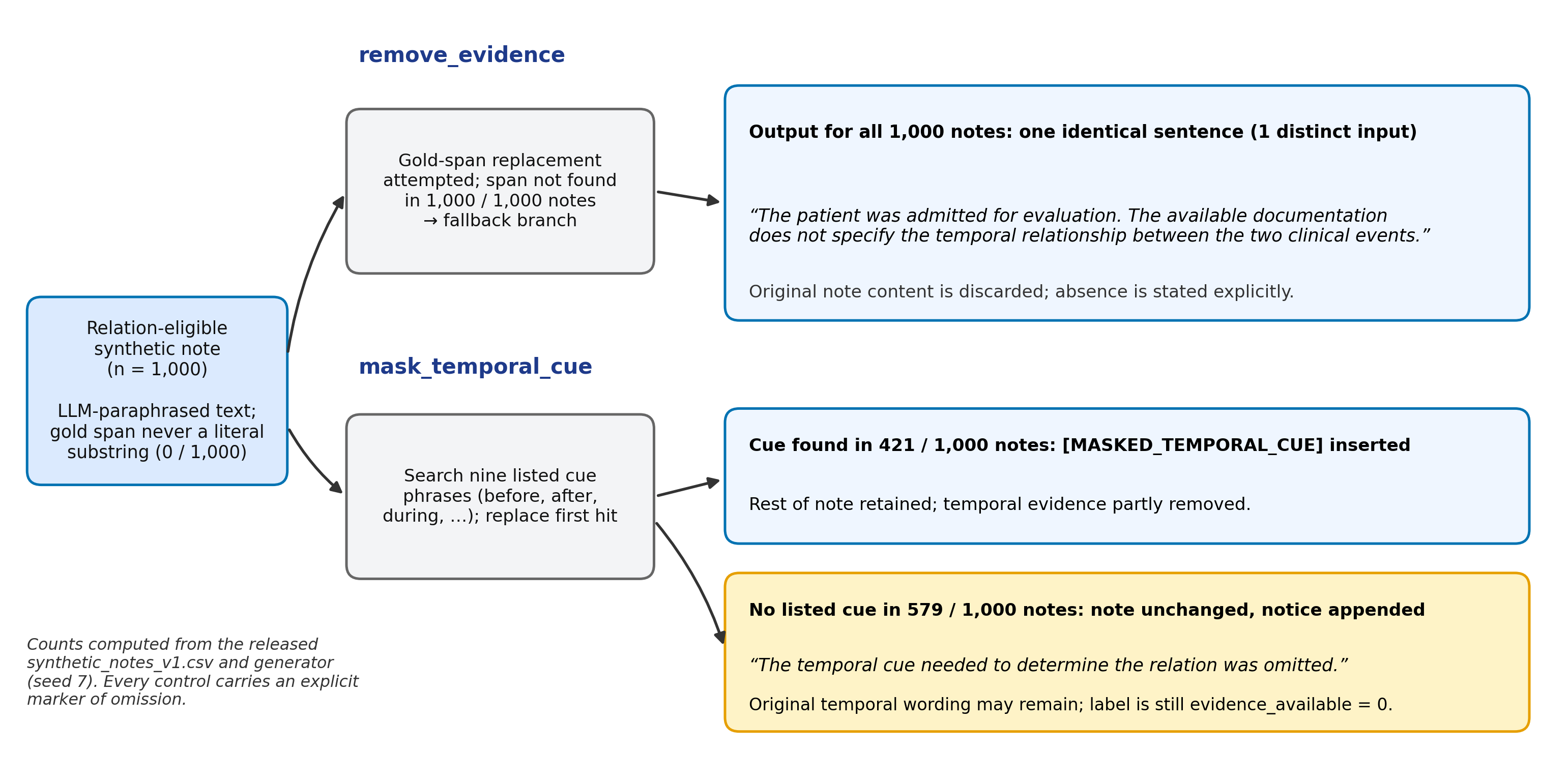}
\caption{Construction and measured composition of the evidence-unavailable controls. \texttt{remove\_evidence} produced one identical sentence for all 1,000 notes. \texttt{mask\_temporal\_cue} masked a cue in 421 notes and, in 579, kept the original note and appended a sentence stating that the cue was omitted.}
\label{fig:controls}
\end{figure}

The control generator reads the 1,000 relation-eligible notes and writes two matched variants of each, both labelled \texttt{evidence\_available\_oracle} $=0$ and \texttt{contradiction\_present\_oracle} $=0$.
\begin{itemize}[nosep]
\item \textbf{\texttt{remove\_evidence}}: The gold evidence span is replaced by the sentence ``The note does not document the temporal relationship between the two events.'' If the span is not found in the note, the whole note is replaced by a fixed text: ``The patient was admitted for evaluation. The available documentation does not specify the temporal relationship between the two clinical events.''
\item \textbf{\texttt{mask\_temporal\_cue}}: The nine cue phrases are tried in a fixed order and the first occurrence (case-insensitive) of the first phrase found (\emph{prior to, before, preceding, following, after, subsequent to, during, concurrent with, while undergoing}) is replaced by \texttt{[MASKED\_TEMPORAL\_CUE]}. If no phrase matches, the note is left unchanged and the sentence ``The temporal cue needed to determine the relation was omitted.'' is appended.
\end{itemize}
Gold spans were recorded before paraphrasing and none of the 1,000 gold spans occurs as a literal substring of its note (the same defect that zeroes evidence recall, Section~\ref{sec:span}). The fallback branch was therefore taken for all 1,000 \texttt{remove\_evidence} variants. For \texttt{mask\_temporal\_cue}, a listed cue was found and masked in 421 notes; in the other 579 the paraphrased wording used none of the nine phrases, so the note was kept and the omission sentence appended. Figure~\ref{fig:controls} shows the construction. These counts were computed by running the released generator (seed 7) on the released \texttt{synthetic\_notes\_v1.csv}.

Four properties of the control set follow and are carried through the rest of the paper. (i) \texttt{remove\_evidence} is a single distinct input repeated 1,000 times (150 times in the BERT test partition). (ii) \texttt{mask\_temporal\_cue} is a genuine cue masking in 42.1\% of notes and a retained note plus an omission notice in 57.9\%. (iii) Every control carries an explicit textual marker of omission. (iv) The oracle label describes the intervention, not verified absence of inferable temporal information; in the 579 retained notes the original temporal wording may still be readable. The generator also defines two contradiction variants (\texttt{contradiction}, \texttt{remove\_and\_contradict}); they were not evaluated here.

\subsection{Silent-removal and silent-keep controls}\label{sec:silent}
Because every control in Section~\ref{sec:controls} carries an explicit marker, a gate could abstain by detecting the marker rather than the absence of evidence. A second control set removes the marker. For each relation-eligible note that passed the pre-check described below, an LLM editor (the same Mistral-Small-3.2-24B model used as the LLM gate~\cite{mistral2025}) was instructed to delete the temporal relationship between the diagnosis (Event 1) and the treatment initiation (Event 2) while introducing no statement of omission, placeholder, bracket or notice and to always mention Event 1 before Event 2 so that mention order could not carry the answer. A separate judge call to the same model, using the candidate edit alone (the same information an extraction gate would have), then checked two things: that no explicit temporal cue remained (day numbers, dates, ordinals, temporal connectives, pre/post-operative phrasing, checked by regular expression) and that the judge could not determine the gold relation from the edit. An edit meeting both conditions was accepted as \texttt{silent\_removal}. When the judge could still infer an order from residual phrasing (for example causal or purpose language) after all explicit cues were removed, the edit was retained as a weaker \texttt{implied\_only} tier rather than discarded, so that this residual-inference case is visible rather than merged with the clean removals (\texttt{strict} tier). Up to five attempts were made per note, with the judge's stated evidence fed back to the editor on retry.

A matched negative control, \texttt{silent\_keep}, was built with the same editor and judge but a different instruction: a minimal synonym-level edit keeping structure, every day number and the temporal relationship (length ratio 0.70–1.40), accepted only if the judge recovered the gold relation. Unlike \texttt{silent\_removal} it imposes no mention order and no list restructuring. A change under \texttt{silent\_removal} but not \texttt{silent\_keep} is therefore attributable to the removal edit as a whole (deleted temporal evidence, imposed Event-1-first order and restructuring), not to deleted evidence alone.

Before any edit, the judge was run on the unedited note, and a note was carried forward only if the judge recovered its gold relation (561 of 1,000). The accepted controls are therefore restricted to notes whose relation the gate's own model family already reads correctly. Of the 561 notes passing this check, 507 produced an accepted \texttt{silent\_removal} edit and 452 an accepted \texttt{silent\_keep} edit; the remainder exhausted five attempts without a validated edit. The BERT test partition (patient-level, seed 7) restricts this to 81 \texttt{silent\_removal} notes (37 strict, 44 implied-only) and 70 \texttt{silent\_keep} notes. The LLM gate, which needs no training split, was evaluated on the full accepted sets: 507 \texttt{silent\_removal} notes (238 strict, 269 implied-only) and 452 \texttt{silent\_keep} notes. Editing and judging used the same model family as the LLM gate under test; a second, independently sourced judge model was not available and is listed as a limitation.

\subsection{Post-paraphrase gold-span regeneration}\label{sec:goldspans}
The gold evidence spans that ground the recall metric in Section~\ref{sec:span} were written before paraphrasing and are not literal substrings of the paraphrased notes, which invalidates exact-match recall. Spans were regenerated directly against the released, paraphrased note text. For each relation-eligible note, an LLM relocator was given the original (pre-paraphrase) gold span and the paraphrased note and asked to copy, verbatim, the contiguous substring of the paraphrased note expressing the same temporal fact, or to return null if the information did not survive paraphrasing. A candidate was accepted only if it was an exact substring of the note and a separate judge call to the same model, given the candidate span alone (mirroring what an extraction gate is given as ``cited evidence''), confirmed it entailed the note's gold relation; a further check for whether the candidate conveyed the same information as the original pre-paraphrase span split accepted spans into \texttt{verified} and \texttt{relation\_only} tiers. Up to three attempts were made per note, with the judge's evidence fed back on retry.

Of 1,000 notes, 482 produced an accepted span (466 \texttt{verified}, 16 \texttt{relation\_only}); 139 were judged to have lost the temporal information during paraphrasing (\texttt{evidence\_lost\_in\_}\allowbreak\texttt{paraphrase}); the remaining 379 exhausted three attempts without a validated relocation. As in Section~\ref{sec:silent}, the same model family served as both relocator and judge. Because relocator, judge and LLM gate are the same model family, agreement between the gate's cited span and the regenerated span partly reflects agreement of one model with itself.

Recall against the regenerated spans is computed differently for the two gates, reflecting what each gate outputs. The LLM cites a span of text directly, so recall is a character-overlap fraction between the cited span and the regenerated gold span, both located in the note; precision is the complementary fraction of the cited span that falls inside the gold span and the cited span's length as a fraction of the note length is reported alongside to distinguish genuine localization from a span that simply covers most of the note. BERT does not output span text; its evidence head produces a per-token score. Recall is therefore a token-overlap fraction: gold tokens are those falling inside the regenerated span's character range and predicted tokens are those with a score at or above 0.1, the same threshold used elsewhere for token-mask coverage; predicted span text, where reported, is reconstructed as the longest contiguous run of tokens meeting that threshold. BERT's token-overlap recall and the LLM's character-overlap recall are not on a comparable scale and no LLM-style precision was computed for BERT.

\subsection{Evidence gates}
\label{sec:evidence_gates}
\paragraph{BioClinicalBERT, hard-gated.} A linear evidence-scoring head and a relation classifier were trained on a frozen BioClinicalBERT encoder. Training combined a prediction loss, a coverage-floor hinge loss (penalizing coverage below a target, added to prevent the evidence mask from collapsing toward zero), sufficiency and necessity losses (predictions from evidence tokens only versus non-evidence tokens only) and a causal-consistency loss (Spearman correlation between per-token evidence scores and their leave-one-out ablation effect). At inference the hard gate abstains when the token-mask coverage falls below a floor. A parallel ungated model shared the architecture and training but did not enforce the gate.

\paragraph{BioClinicalBERT, availability-supervised.} This variant keeps the frozen BioClinicalBERT encoder and adds a third linear head that predicts whether the note contains the evidence. The evidence head scores each token and the scores weight a pooled representation from which the relation and availability logits are computed. Training minimises the relation cross-entropy plus 0.3 times a token-level binary cross-entropy for evidence plus 1.0 times a binary cross-entropy for availability (AdamW, learning rate $5\times10^{-5}$, batch size 8, 10 epochs, maximum length 128 tokens, seed 7). Each training-partition clean note (availability 1) is accompanied by its \texttt{remove\_evidence} and \texttt{mask\_temporal\_cue} controls (availability 0), built by the same functions that build the test controls. At inference the gate abstains when the predicted availability probability is below 0.5. The evidence labels come from gold spans recorded before paraphrasing, which are not substrings of the notes, so they are all zero and the evidence head receives no informative supervision (Section~\ref{sec:span}). The hard-gated model derives its evidence labels the same way. Per-token evidence scores of both BERT models, and the recall computed from them, therefore do not indicate what a span-supervised BERT head would locate. Inputs are truncated at 128 tokens; about 5\% of relation-eligible notes exceed roughly 90 words, so an appended notice can fall outside BERT's input. The development split was not used for model selection. 

\paragraph{Zero-shot LLM.} LLM-based temporal entity and relation extraction from clinical reports has been shown to work in a low-resource setting, with Mistral the strongest of the models compared \cite{andrew2025jamia, andrew2024cl4h}; Mistral is used as the zero-shot gate. The LLM returns JSON with a relation (or null), a verbatim evidence span, a confidence and a reason, at temperature 0.0 through an OpenAI-compatible endpoint. The gate abstains when (a) the relation is null (\texttt{no\_relation}) (b) the cited span is not a substring of the note, is shorter than 10 characters, or covers under 3\% of the note (\texttt{insufficient\_evidence\_span}) or (c) fewer than half of up to three re-queries with evidence-focused wording return the same relation (consistency check failed; recorded as \texttt{necessity\_check\_failed} in the output). Unlike the original pipeline (Table \ref{tab:orig}), which masked the cited span before re-querying and so tested necessity, these re-queries leave the note unchanged and test stability under rewording only. The abstention and recall experiments use one configuration: an availability-aware prompt that defines EVENT1 as the diagnosis and EVENT2 as the treatment initiation, gives the three relation definitions and allows a null relation and null evidence span when ``the note does not contain clear temporal evidence''; a minimum span of 10 characters and 3\% of the note; three consistency re-queries; and an agreement threshold of 0.5. Clean predictions, all control predictions and the cited evidence spans used for recall were produced with the same script and settings (\texttt{llm\_availability\_gated\_inference\_v2.py}, \texttt{--event\_mode convention}). An earlier run of the same gate without the event definitions (\texttt{\_v1.py}) is reported once in Table~\ref{tab:clean} for comparison; without the definitions the model had to infer which event was which and its answered accuracy was close to the majority-class rate. The gate reads no oracle label; the script copies oracle columns into its output for bookkeeping only. The original-gate results in Table~\ref{tab:orig} come from the earlier LLM pipeline.

BERT coverage is a token-mask fraction and LLM coverage is a character-span fraction; the two are not comparable.

\subsection{Evidence Graph compression operators}
An uncompressed Evidence Graph was built in which every (fact, note) mention is a distinct node carrying a simulated per-mention extraction confidence. Three operators were applied in sequence.
\begin{itemize}[nosep]
\item \emph{Aggregation} merges all mentions of a fact into one state node, keeping membership edges to every constituent mention (reversible by construction).
\item \emph{Revision-representation} collapses a fact's revision chain (state nodes linked by \emph{revises} edges) into one evidential-state node with an ordered revision history, replacing $N$ nodes and $N-1$ edges with one node.
\item \emph{Temporal-approximation} removes the exact day from any state or revision-history entry whose simulated confidence is below 0.65 and records a temporal-relation edge to a same-patient, same-concept anchor day~\cite{allen1983}. The operator is intentionally lossy: an approximate relation may leave a query unresolved, in which case the representation returns ABSTAIN.
\end{itemize}
A fixed-window baseline (keep the $k=2$ most recent mentions per concept) isolates how much of any advantage comes from provenance-aware compression rather than compression in general. Every representation, including the uncompressed one, is scored on representation size (nodes plus edges), query accuracy against the revision-sensitive oracle and provenance recall. Query coverage is the fraction of sampled query points for which the representation returns a resolved answer rather than ABSTAIN. Six query days were sampled per patient and concept type.

Size counts nodes plus edges. Aggregation moves each mention's identifier, day and confidence from a node onto its membership edge, reducing the count without discarding mention information, so serialized size (JSON bytes) is also reported. The query returns the most recent entry dated on or before the query day and does not traverse revises edges; because every revision post-dates the fact it revises, a revision-blind ``latest fact by day'' rule matches the oracle at all 6,813 points, so the experiment tests whether compression preserves the information needed to answer, not revision reasoning. Only 958 points (14.1\%) come from the 110 patients with a revision and 495 fall after a revision; results are reported for these subsets too. A representation holding no entry on or before the query day returns no value, scored as incorrect and, unlike ABSTAIN, counted as covered.
\subsection{Abstention evaluation protocol}
For each gate, clean predictions on evidence-available notes and control predictions on evidence-unavailable inputs were scored by the evaluation script. The BERT gate was evaluated on the 150-note test partition (patient-level split, seed 7) and the 300 controls derived from the same patients. The LLM gate needs no training, so it was evaluated on all 1,000 relation-eligible notes and all 2,000 controls. Let $n$ be the number of inputs in a condition, $A$ the number answered and $A_c$ the number answered with the gold relation.

On clean inputs: coverage $=A/n$; false-abstention rate $=1-A/n$; answered accuracy $=A_c/A$. On evidence-unavailable controls: abstention recall $=1-A/n$; coverage on unavailable evidence $=A/n$; unsafe-answer rate $=(A-A_c)/n$; answered accuracy $=A_c/A$, which is undefined when $A=0$. The unsafe-answer rate counts answered inputs whose relation differs from the gold relation of the original note. It is therefore not the same as coverage on unavailable evidence: a correct answer on a control is not counted as unsafe, even though the oracle says no evidence was available. Both quantities are reported. Controls were assigned to variants by the \texttt{evidence\_case} field.

Wilson 95\% intervals~\cite{wilson1927} are reported for the abstention metrics and treat inputs as independent. Clean and control inputs share base notes, so they are paired. Intervals are not reported for \texttt{remove\_evidence}, which is a single distinct input. The \texttt{mask\_temporal\_cue} results were also split by the control file's \texttt{masked\_cue} field and each control was paired with the clean prediction for its base note (\texttt{stratify\_masked\_cue.py}); neither needs additional inference. Because the BERT gate is trained and the LLM gate is zero-shot and because sample sizes and evidence representations differ, BERT and LLM values are descriptive and are not ranked.

\subsection{Temporally versioned provenance graph}
The provenance graph is model-agnostic infrastructure that ingests either extractor's per-note gate output into a structure separate from the compression graph. Four properties are exercised. (1) \emph{Entity resolution:} nodes are keyed by (note id, label, scale), so re-run extractions of the same event resolve to the same node. (2) \emph{Never-overwrite versioning:} a new extraction of a known relation creates a new edge; disagreeing edges are cross-flagged (\emph{contradicts}/\emph{contradicted by}); a reviewer can close an edge's validity window and the full history and any point-in-time snapshot are queryable. (3) \emph{Provenance-first storage:} every edge stores the gate's own diagnostics verbatim (coverage, necessity outcome, evidence spans and indices, abstention reason, model identifier) and abstentions are stored as an explicit ABSTAINED edge type rather than dropped. (4) \emph{Federation-ready aggregation:} only summary statistics (relation-type counts, mean coverage, abstention rate) are computed, so they can cross a site boundary without raw text or spans. In the reported experiment the graph compared ungated and gated LLM outputs on the same notes. Direct BERT-versus-LLM contradiction evaluation was not performed.

\subsection{Secondary analyses: perturbation, manipulation and certification}
For each of 30 sampled BERT or 20 sampled LLM clean examples, 100 perturbed variants were generated by abbreviation substitution, OCR-like character changes, token insertion or synonym substitution, with an edit budget of 1--2 edits. A one-sided Clopper--Pearson lower bound~\cite{clopper1934} was computed separately for relation stability and evidence stability as \texttt{scipy.stats.beta.ppf(alpha/2, successes, n - successes + 1)}, with the significance level split Bonferroni-style across the two. An example was labelled CERTIFIED only when both bounds reached $1-\alpha$. The BERT and LLM runs used different $\alpha$ and different subsets, so their certification rates are descriptive. Manipulation analyses used paired base and adversarial variants; an example counts as stable only when both calls answer and return the same relation. Perturbation-level observations were not treated as independent patient-level samples.

\section{Results}

\subsection{Corpus}
The generator reported 3,353 completed paraphrase calls; this counts successful returns and does not establish that facts and temporal relations were preserved.

\begin{table}[t]
\centering\small
\caption{Dataset summary.}
\label{tab:corpus}
\begin{tabular}{@{}p{6.4cm}p{8.6cm}@{}}
\toprule
Quantity & Value \\
\midrule
Patients & 1,000 \\
Notes (total) & 3,353 (realized mean 3.35 per patient) \\
Clinical facts & 2,220 \\
Patients with a documented revision & 110 (11.0\%) \\
Revision edges (ground truth) & 220 \\
Relation-eligible notes (gold relation and span) & 1,000 \\
Adversarial (contradiction-injected) variants & 2,000 \\
Evidence-unavailable controls & 2,000 (1,000 \texttt{remove\_evidence}, 1,000 \texttt{mask\_temporal\_cue}) \\
Note styles & terse ICU 816, verbose discharge 844, structured admission 857, nursing 836 \\
LLM paraphrase success & 3,353 / 3,353 (100\%) \\
\bottomrule
\end{tabular}
\end{table}

\subsection{Evidence Graph compression}
\begin{figure}[t]
\centering
\includegraphics[width=\textwidth]{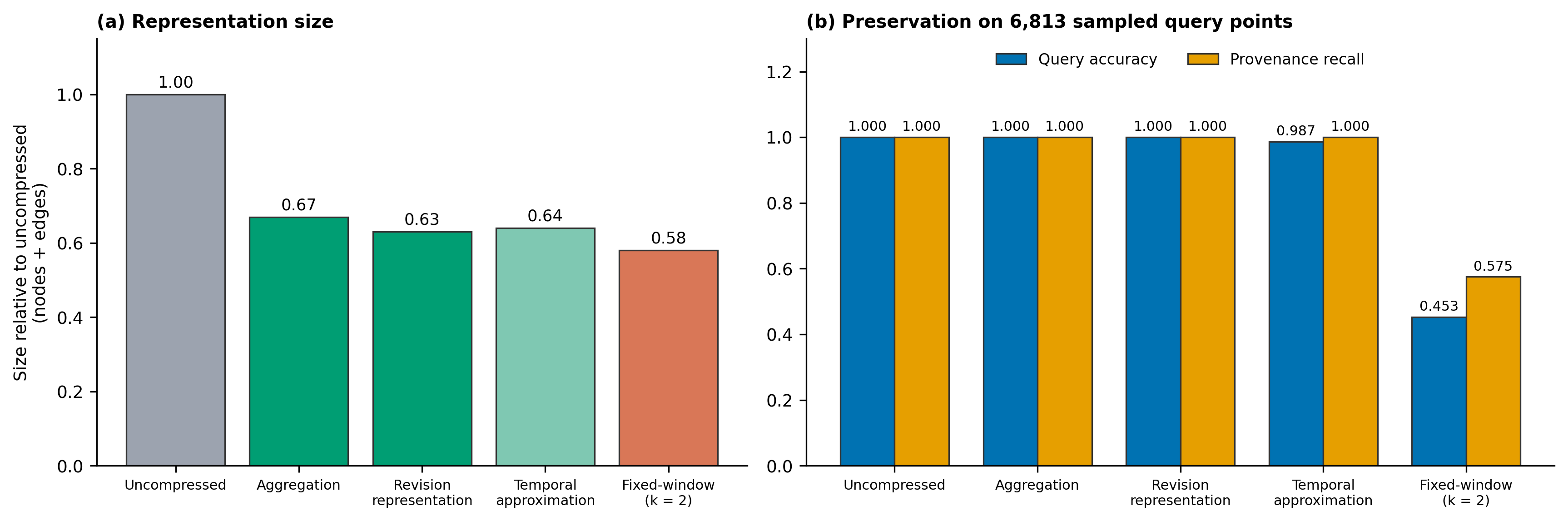}
\caption{Compression trade-off. (a) Representation size relative to the uncompressed graph (nodes plus edges). (b) Query accuracy and provenance recall on the same 6,813 sampled query points. The fixed-window baseline is the smallest representation and the only one that loses mention links. Temporal-approximation returns unflagged wrong values on 1.3\% of points; the fixed window returns no value on 53.4\% and a wrong value on 1.2\%, none flagged.}
\label{fig:compression}
\end{figure}

Table~\ref{tab:compression} and Figure~\ref{fig:compression} summarize the compression experiment over $n$ = 6,813 sampled (patient, concept, query-day) points across 1,000 patients. Aggregation reduced total size to 0.67 of the uncompressed representation and revision-representation to 0.63. Both retained query accuracy 1.000 and provenance recall 1.000. These are preservation results for the sampled query set and are consistent with the structural reversibility of the operators.

\begin{table}[h]
\caption{Representation comparison across compression strategies (6,813 sampled query points).}\label{tab:compression}%
\footnotesize
\setlength{\tabcolsep}{4pt}
\begin{tabular}{@{}lrrrrrrr@{}}
\toprule
\multicolumn{8}{@{}l}{\textit{A. All query points ($n = 6{,}813$)}}\\
Representation & Nodes & Edges & Total & Size ratio & Query acc. & Query cov. & Prov. recall\\
\midrule
Uncompressed            & 6,926 & 7,146 & 14,072 & 1.00 & 1.000 & 1.000 & 1.000\\
Aggregation             & 2,220 & 7,146 &  9,366 & 0.67 & 1.000 & 1.000 & 1.000\\
Revision-representation & 2,000 & 6,926 &  8,926 & 0.63 & 1.000 & 1.000 & 1.000\\
Temporal-approximation  & 2,000 & 6,978 &  8,978 & 0.64 & 0.987 & 0.996 & 1.000\\
Fixed-window ($k = 2$)  & 3,982 & 4,202 &  8,184 & 0.58 & 0.453 & 1.000 & 0.575\\
\midrule
\multicolumn{8}{@{}l}{\textit{B. Serialized size and revision subsets}}\\
 & Bytes & \multicolumn{4}{c}{Revision patients ($n = 958$)} & \multicolumn{2}{c@{}}{Post-revision ($n = 495$)}\\
Representation & ratio & Acc. & Cov. & No value & Wrong & Acc. & No value / Wrong\\
\midrule
Uncompressed            & 1.00 & 1.000 & 1.000 &   0 &  0 & 1.000 & 0 / 0\\
Aggregation             & 0.77 & 1.000 & 1.000 &   0 &  0 & 1.000 & 0 / 0\\
Revision-representation & 0.78 & 1.000 & 1.000 &   0 &  0 & 1.000 & 0 / 0\\
Temporal-approximation  & 0.78 & 0.907 & 0.974 &   0 & 87 & 0.824 & 0 / 87\\
Fixed-window ($k = 2$)  & 0.58 & 0.401 & 1.000 & 490 & 84 & 0.776 & 27 / 84\\
\botrule
\end{tabular}
\footnotetext{Size ratio is nodes plus edges divided by the uncompressed total; bytes ratio is the same for the JSON-serialized representation. Query acc. is accuracy over answered (non-ABSTAIN) points and Query cov. the fraction answered. A representation with no entry dated on or before the query day returns no value, which is scored as wrong and counted as covered. Over all points, temporal-approximation answered 6,788 (6,701 correct, 87 wrong without a flag, 1.3\%) and returned ABSTAIN on 25 (0.4\%); the fixed window returned no value on 3,641 (53.4\%) and a wrong value on 84 (1.2\%). Revision patients: the 110 patients with a revision edge. Post-revision: points on or after the day a revision of the queried concept took effect; temporal-approximation returned no ABSTAIN there. Because every revision post-dates the fact it revises, a revision-blind latest-fact-by-day rule scores 1.000 on all three sets.}
\end{table}

The fixed-window baseline was smaller (0.58) but reached 0.453 query accuracy and 0.575 provenance recall. Most of its failures (3,641 of 3,725) were empty returns: a mention carries its note's day, and the two latest mentions often post-date the query day. Only 84 (1.2\%) were wrong values, and neither failure type was flagged. With k = 3 and 4 it reached 0.762 and 0.941 accuracy at 0.82 and 0.95 of the size. On the 495 post-revision points it reached 0.776, against 0.824 for temporal-approximation and 1.000 for the lossless operators.

Two features of the size figures matter for auditability. Aggregation removed 4,706 nodes but no edges (7,146 before and after), because a membership edge is kept for every mention; the saving is bounded by the provenance links the design deliberately retains. Measured in serialized bytes, the saving is smaller (0.77). Temporal-approximation is the intentionally lossy operator. It kept 2,000 nodes, added 52 temporal-relation edges (6,978 versus 6,926) and reached 0.987 accuracy on the queries it answered, with 0.996 coverage. Of the 6,813 query points, 25 (0.4\%) were returned as ABSTAIN and 87 (1.3\%) were answered wrongly without a flag, so the explicit ABSTAIN caught 25 of the 112 queries the operator did not answer correctly. It is larger than revision-representation because of these 52 edges and buys an explicit approximation record, not further compression. All 112 errors and abstentions occurred in patients with a revision (11.7\% of their 958 points). Because confidence values were simulated rather than produced by the extraction models, the experiment evaluates the operators under the generator's confidence model and not under calibrated model confidence.

\subsection{Relation extraction under the original gates and the evidence-span recall defect}\label{sec:span}
Table~\ref{tab:orig} gives the original gate results. On the 150-note BERT test set the ungated model answered every example with accuracy 0.740. The hard-gated model abstained on 144 of 150 examples (0.960); its accuracy of 1.000 rests on six answered examples and is not a stable estimate. Its evidence head was trained toward all-zero labels (Section \ref{sec:evidence_gates}) and the coverage floor is applied to those scores, so this abstention rate is best read as a consequence of the span defect rather than of the gating design. On the 60-note LLM subset (drawn from the same 150-patient held-out split) the baseline reached 0.617 and the gated pipeline abstained on 55.0\% of inputs with accuracy 0.556 on the 27 it answered. The rows are descriptive: they differ in sample size and evidence representation.

\begin{table}[t]
\centering\small
\caption{Relation extraction under the original gates. Evidence recall is invalid as implemented (see text).}
\label{tab:orig}
\begin{tabular}{@{}llrrrr@{}}
\toprule
System & Configuration & $n$ & Abstain rate & Accuracy (answered) & Evidence recall \\
\midrule
BERT & Baseline (ungated) & 150 & 0.000 & 0.740 & 0.000 \\
BERT & Provenance-first (hard-gated) & 150 & 0.960 & 1.000 (6 answered) & 0.000 \\
LLM & Baseline (ungated) & 60 & 0.000 & 0.617 & 0.000 \\
LLM & Provenance-first (gated) & 60 & 0.550 & 0.556 (27 answered) & 0.000 \\
\bottomrule
\end{tabular}
\end{table}

Mean evidence-span recall was 0.000 in all four configurations. The metric is scored only for examples whose gold span matches the note text literally and defaults to 0.0 when none do. Gold spans were written before paraphrasing and none of the 1,000 relation-eligible notes contains its gold span as a literal substring. The value is therefore a fallback default. It says nothing about whether either architecture located the supporting text and it cannot rank gates on this corpus. Section~\ref{sec:recall2} reports recall against the regenerated spans of Section~\ref{sec:goldspans}, which does not have this defect.

\subsection{Evidence-span recall against regenerated gold spans}\label{sec:recall2}
\begin{figure}[t]
\centering
\includegraphics[width=\textwidth]{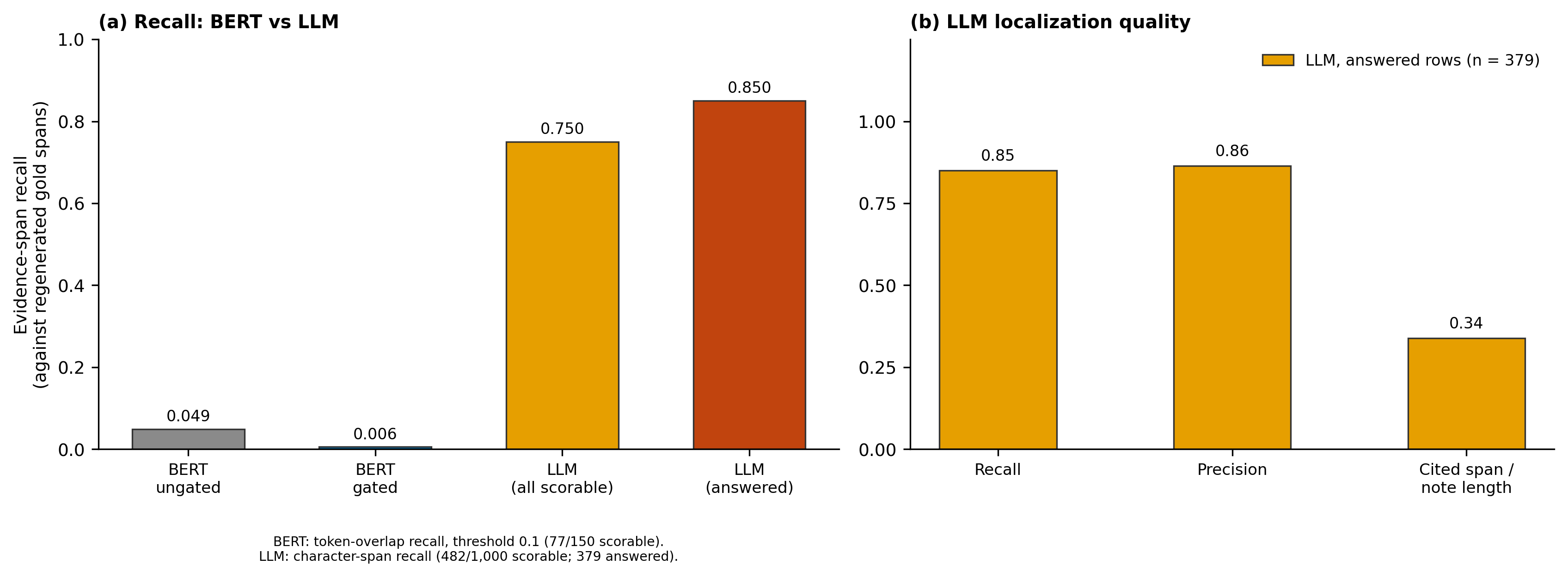}
\caption{Evidence-span recall against the 482 regenerated gold spans of Section~\ref{sec:goldspans}. (a) Recall for the ungated and availability-supervised (gated) BERT models (token-overlap, 77/150 scorable) and the LLM gate, pooled (abstained rows scored on the span they cited, 0 if none) and restricted to answered rows. (b) LLM recall, precision and the cited span's length as a fraction of the note, on answered rows: high recall and precision together with a short span show that the gate did not simply cite most of the note.}
\label{fig:recall2}
\end{figure}

Table~\ref{tab:recall2} and Figure~\ref{fig:recall2} give recall, and, for the LLM, precision and cited-span length, against the 482 regenerated gold spans of Section~\ref{sec:goldspans}. The LLM spans come from a second run of the same gate configuration, made to export the cited span for each row (\texttt{llm\_v2\_spans}). It answered 694 of 1,000 clean notes (accuracy 0.663), against 700 (accuracy 0.663) in the run used for Tables~\ref{tab:clean}--\ref{tab:silent}; the six-note difference reflects non-deterministic inference at temperature 0.0.

\begin{table}[t]
\centering\small
\caption{Evidence-span recall against regenerated gold spans. LLM precision is the fraction of the cited span that falls inside the gold span; span/note length is the cited span's length as a fraction of the note's length. BERT reports token-overlap recall only (Section~\ref{sec:goldspans}); no LLM-style precision was computed for it. BERT heads trained on all-zero span labels.}
\label{tab:recall2}
\begin{tabular}{@{}llrlll@{}}
\toprule
Gate & Rows & $n$ scorable & Recall & Precision & Span / note length \\
\midrule
BERT & Ungated baseline & 77 / 150 & 0.049 & --- & --- \\
BERT & Availability-supervised (gated) & 77 / 150 & 0.006 & --- & --- \\
LLM & All scorable (abstentions scored on their cited span) & 482 / 1{,}000 & 0.750 & 0.866 & 0.329 \\
LLM & Answered rows only & 379 / 1{,}000 & 0.850 & 0.864 & 0.339 \\
\bottomrule
\end{tabular}
\end{table}

Because both BERT evidence heads were trained without span supervision, these values describe an unsupervised head, not BERT's capacity to localize evidence. On the 77 of 150 BERT test notes with a regenerated gold span, the ungated model's recall was 0.049 and the availability-supervised (gated) model's was 0.006, roughly ninefold lower. Inspection of the ungated model's reconstructed spans showed many single-token or sub-word fragments (for example the isolated word-piece \texttt{rrel}, from within ``correlating'') and generic connective phrases the paraphraser tends to insert near symptom mentions (\texttt{correlating with persistent}, \texttt{manage associated}), rather than dates or temporal connectives. Whether the gated model's reconstructed spans show the same pattern was not examined and is listed as a limitation.

On the 482 LLM rows with a regenerated gold span, mean recall was 0.750 (median 1.000; 53.3\% of rows at 1.000; 11.4\% at exactly 0). Restricting to the 379 rows the gate answered (most abstentions still cited a span, which the pooled figure scores; abstentions without a span score 0) gives recall 0.850 and precision 0.864. The cited span covered a mean of 32.9\% (answered rows: 33.9\%) of the note and no scorable row had a cited span covering more than 80\% of its note, so the recall is not attributable to citing most of the note. The notes are short, however (median 33 words), and the pre-paraphrase evidence sentence was a median 69\% of note length, so covering a third of the note is a modest localization requirement. Mean recall was 0.747 on the verified tier and 0.849 on the relation only tier; the latter has only 16 rows, so the difference is not interpretable.

\subsection{Evidence-availability abstention}\label{sec:abstain}
\begin{figure}[t]
\centering
\includegraphics[width=\textwidth]{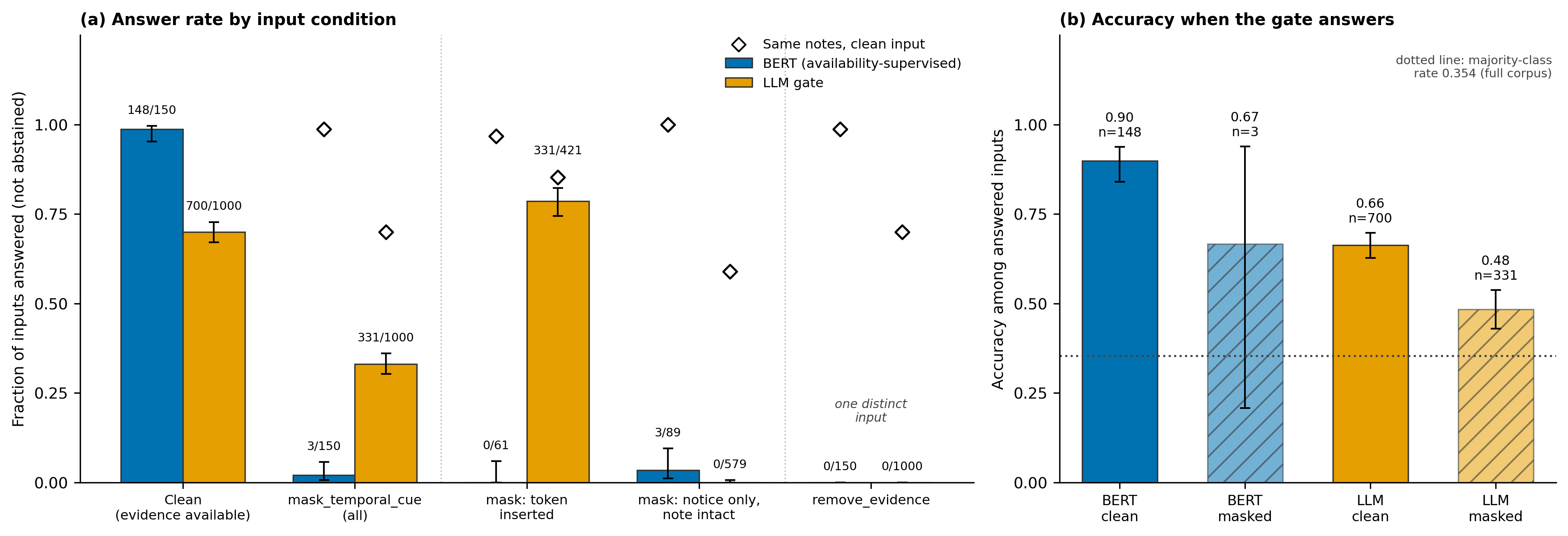}
\caption{Abstention behaviour of the availability-supervised BERT gate and the LLM gate. (a) Fraction of inputs answered, with Wilson 95\% intervals; an ideal gate answers clean inputs and abstains on every control. Diamonds mark the answer rate of the same base notes when clean, so the distance from a diamond to its bar is the effect of the control. The two middle conditions split \texttt{mask\_temporal\_cue} by whether a cue was replaced by a mask token or the note was left intact with an appended notice. \texttt{remove\_evidence} is a single distinct input. (b) Accuracy among answered inputs; hatched bars are \texttt{mask\_temporal\_cue}, scored against the gold relation of the original note. The dotted line is the majority-class rate.}
\label{fig:abstain}
\end{figure}

Table~\ref{tab:clean} shows clean-input behaviour and Table~\ref{tab:control} shows behaviour on the controls; Figure~\ref{fig:abstain} plots both. Two points anchor the reading of accuracy: the majority-class rate is 0.354 and the availability-supervised BERT is a separately trained model, so its answered accuracy is not comparable with the 0.740 of the ungated baseline.

\begin{table}[t]
\centering\scriptsize
\setlength{\tabcolsep}{4pt}
\caption{Clean (evidence-available) inputs. Wilson 95\% intervals in brackets for the new runs; the false-abstention rate is 1 minus coverage and shares its interval. Prompt: \emph{original} is the earlier LLM pipeline of Table~\ref{tab:orig}; \emph{event defs.}\ means EVENT1/EVENT2 and the three relations are defined in the prompt. All other LLM results in this paper use the prompt with event definitions. The hard-gated BERT abstains below a token-coverage floor.}
\label{tab:clean}
\begin{tabular}{@{}lllrlrl@{}}
\toprule
System & Gate & Prompt & $n$ & Coverage & False abst. & Answered acc. ($n$ ans.) \\
\midrule
BERT & Ungated baseline & --- & 150 & 1.000 & 0.000 & 0.740 (150) \\
BERT & Hard-gated & --- & 150 & 0.040 & 0.960 & 1.000 (6) \\
BERT & Availability-supervised & --- & 150 & 0.987 [0.953, 0.996] & 0.013 & 0.899 [0.840, 0.938] (148) \\
LLM & Ungated baseline & original & 60 & 1.000 & 0.000 & 0.617 (60) \\
LLM & Original gate & original & 60 & 0.450 & 0.550 & 0.556 (27) \\
LLM & Availability-aware & no event defs. & 1,000 & 0.587 [0.556, 0.617] & 0.413 & 0.373 [0.335, 0.413] (587) \\
LLM & Availability-aware & event defs. & 1,000 & 0.700 [0.671, 0.728] & 0.300 & 0.663 [0.627, 0.697] (700) \\
\bottomrule
\end{tabular}
\end{table}

\begin{table}[t]
\centering\scriptsize
\setlength{\tabcolsep}{3pt}
\caption{Evidence-unavailable controls, with each control's paired clean coverage. Clean cov.\ is the answer rate of the same base notes without any intervention. Coverage on unavailable evidence equals 1 minus abstention recall (for example 0.331 for the LLM on \texttt{mask\_temporal\_cue}) and shares its interval. Wilson 95\% intervals in brackets; none is given for \texttt{remove\_evidence}, a single distinct input. The indented rows split \texttt{mask\_temporal\_cue} by the control file's \texttt{masked\_cue} field: a mask token replaced a cue phrase, or the note was left intact with an appended omission notice. Answered accuracy is undefined when no input is answered.}
\label{tab:control}
\begin{tabular}{@{}llrrllrl@{}}
\toprule
Gate & Condition & $n$ & Clean cov. & Abstention recall & Unsafe-answer rate & $n$ ans. & Answered accuracy \\
\midrule
BERT & \texttt{remove\_evidence} & 150 & 0.987 & 1.000 & 0.000 & 0 & undefined \\
BERT & \texttt{mask\_temporal\_cue} & 150 & 0.987 & 0.980 [0.943, 0.993] & 0.007 [0.001, 0.037] & 3 & 0.667 \\
BERT & \quad token inserted & 61 & 0.967 & 1.000 [0.941, 1.000] & 0.000 [0.000, 0.059] & 0 & undefined \\
BERT & \quad notice, note intact & 89 & 1.000 & 0.966 [0.906, 0.988] & 0.011 [0.002, 0.061] & 3 & 0.667 \\
BERT & All controls & 300 & 0.987 & 0.990 [0.971, 0.997] & 0.003 [0.001, 0.019] & 3 & 0.667 \\
LLM & \texttt{remove\_evidence} & 1,000 & 0.700 & 1.000 & 0.000 & 0 & undefined \\
LLM & \texttt{mask\_temporal\_cue} & 1,000 & 0.700 & 0.669 [0.639, 0.697] & 0.171 [0.149, 0.196] & 331 & 0.483 [0.430, 0.537] \\
LLM & \quad token inserted & 421 & 0.853 & 0.214 [0.177, 0.255] & 0.406 [0.360, 0.454] & 331 & 0.483 [0.430, 0.537] \\
LLM & \quad notice, note intact & 579 & 0.589 & 1.000 [0.993, 1.000] & 0.000 [0.000, 0.007] & 0 & undefined \\
LLM & All controls & 2,000 & 0.700 & 0.835 [0.818, 0.850] & 0.086 [0.074, 0.099] & 331 & 0.483 [0.430, 0.537] \\
\bottomrule
\end{tabular}
\end{table}

\paragraph{BERT.} The availability-supervised gate answered 148 of 150 clean notes (98.7\%) with accuracy 0.899 and abstained on 297 of 300 controls (99.0\%). Abstention was complete on \texttt{remove\_evidence} and 98.0\% on \texttt{mask\_temporal\_cue}. Three controls were answered, all of them notes carrying the omission notice and two of those answers matched the gold relation. The unsafe-answer rate was 0.33\%. Against the hard-gated model, which abstained on 96.0\% of clean notes, availability supervision removed the over-abstention.

\paragraph{LLM.} The LLM gate answered 70.0\% of clean notes, with accuracy 0.663. It abstained on all 1,000 \texttt{remove\_evidence} inputs and on 66.9\% of \texttt{mask\_temporal\_cue} inputs, answering 331. Those answers were correct in 48.3\% of cases and the unsafe-answer rate was 17.1\% in that condition (8.6\% over all controls). Coverage fell by 36.9 percentage points from clean to masked inputs for the LLM gate and by 96.7 for the BERT gate. The pairing below shows what drove this.

\paragraph{Pairing each control with its clean note.} The control file records the base note of each control and whether a cue was masked, so both gates can be split by control type and compared with their own clean predictions on the same notes, without new inference (Table~\ref{tab:control}, indented rows and clean-coverage column). In the BERT test partition, 61 \texttt{mask\_temporal\_cue} notes carry a mask token and 89 carry the omission notice, as the released generator predicts for the seed-7 split.

For the 89 BERT and 579 LLM notes left intact with an appended notice, one sentence changed the outcome. BERT answered all 89 when clean (accuracy 0.843) and 3 (3.4\%) with the sentence added. The LLM answered 58.9\% of the 579 when clean (accuracy 0.540) and none with the sentence added. BERT's clean accuracy shows that the relation was recoverable from the text of those notes, so the oracle label overstates unavailability there and the abstention is the response to the appended sentence.

Replacing a single cue phrase with a mask token affected the gates differently. BERT answered 96.7\% of the 61 notes when clean (accuracy 0.983) and none with the token. The LLM answered 85.3\% of its 421 notes when clean (accuracy 0.780) and 78.6\% with the token (accuracy 0.483). Its abstention rose only from 14.7\% to 21.4\%, while its accuracy on the notes it answered fell by 29.7 points, so most of the lost evidence went unnoticed; its unsafe-answer rate in this stratum was 40.6\%. BERT's clean accuracy shows only that the relation was recoverable with the cue present. Whether it remains recoverable without it is uncertain, since other temporal wording may survive.

Across the whole control set, 1,579 of the LLM's 1,669 abstentions (94.6\%) came from the two texts that state the omission, the fixed \texttt{remove\_evidence} sentence and the appended notice.

\paragraph{How far these results reach.} Both gates recognised the one explicit statement of absence used in \texttt{remove\_evidence}. That establishes that each gate can abstain on an input that says no temporal information is documented and little beyond it, because 1,000 predictions come from one distinct input. The paired comparisons are the closest Section~\ref{sec:controls}'s marker-carrying controls come to separating the marker from the evidence and they point to the marker for both gates: BERT reacted fully to the notice and to the mask token, the LLM fully to the notice and only weakly to the token. They do not show that either gate abstains when temporal evidence is missing and nothing announces it. Section~\ref{sec:silentresults} reports controls with no marker.

\subsection{Abstention under silent removal}\label{sec:silentresults}
\begin{figure}[t]
\centering
\includegraphics[width=\textwidth]{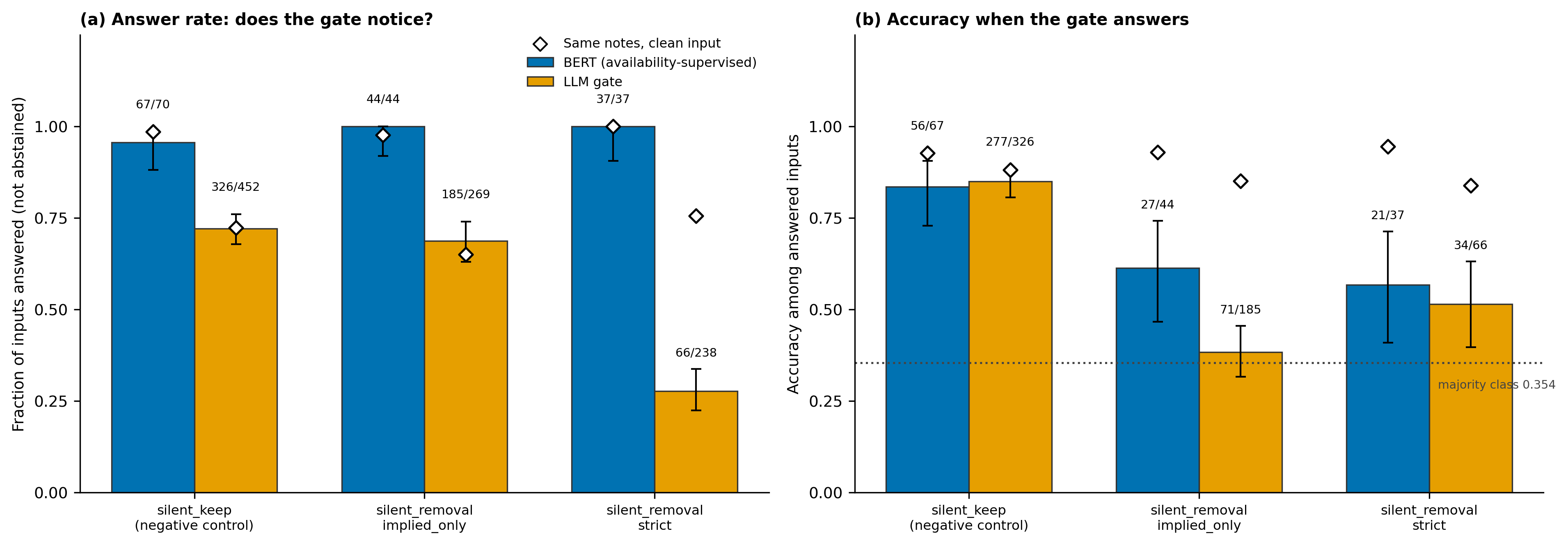}
\caption{Abstention under silent removal. (a) Fraction of inputs answered, with Wilson 95\% intervals; BERT on the seed-7 test partition ($n$ = 70, 44, 37), the LLM on all accepted notes ($n$ = 452, 269, 238). Diamonds mark the answer rate of the same base notes when clean. (b) Accuracy among answered inputs, with diamonds marking accuracy on the same notes when clean; the gap from diamond to bar is the cost of the edit. The dotted line is the majority-class rate.}
\label{fig:silent}
\end{figure}

Table~\ref{tab:silent} and Figure~\ref{fig:silent} report behaviour on the marker-free \texttt{silent\_removal} and \texttt{silent\_keep} controls of Section~\ref{sec:silent}, each paired with the gate's own clean-input behaviour on the same base notes.

\begin{table}[t]
\centering\scriptsize
\setlength{\tabcolsep}{3pt}
\caption{Abstention under silent removal, paired with clean coverage and accuracy on the same base notes. BERT is evaluated on the seed-7 test partition; the LLM, which needs no training split, on all accepted notes. Wilson 95\% intervals in brackets.}
\label{tab:silent}
\begin{tabular}{@{}llrllll@{}}
\toprule
Gate & Condition & $n$ & Clean cov. & Clean acc. & Coverage (silent) & Accuracy (silent) \\
\midrule
BERT & \texttt{silent\_keep} & 70 & 0.986 & 0.928 & 0.957 [0.881, 0.985] & 0.836 [0.729, 0.906] \\
BERT & \texttt{silent\_removal}, implied\_only & 44 & 0.977 & 0.930 & 1.000 [0.920, 1.000] & 0.614 [0.466, 0.743] \\
BERT & \texttt{silent\_removal}, strict & 37 & 1.000 & 0.946 & 1.000 [0.906, 1.000] & 0.568 [0.409, 0.713] \\
BERT & \texttt{silent\_removal}, all & 81 & 0.988 & 0.938 & 1.000 [0.955, 1.000] & 0.593 [0.484, 0.693] \\
LLM & \texttt{silent\_keep} & 452 & 0.723 & 0.881 & 0.721 [0.678, 0.761] & 0.850 [0.807, 0.884] \\
LLM & \texttt{silent\_removal}, implied\_only & 269 & 0.651 & 0.851 & 0.688 [0.630, 0.740] & 0.384 [0.317, 0.456] \\
LLM & \texttt{silent\_removal}, strict & 238 & 0.756 & 0.839 & 0.277 [0.224, 0.337] & 0.515 [0.397, 0.632] \\
LLM & \texttt{silent\_removal}, all & 507 & 0.700 & 0.845 & 0.495 [0.452, 0.538] & 0.418 [0.359, 0.480] \\
\bottomrule
\end{tabular}
\end{table}
\begin{table}[h]
\caption{BERT accuracy under silent removal and silent keep, split by gold relation. Availability-supervised (gated) BERT on the seed-7 test partition; accuracy is over answered notes and Wilson 95\% intervals are in brackets. The gate answered every \texttt{silent\_removal} note. Answered as gives the number of \texttt{silent\_removal} notes assigned each relation.}\label{tab:silent_perclass}%
\footnotesize
\setlength{\tabcolsep}{3.5pt}
\begin{tabular}{@{}lrcccrcc@{}}
\toprule
 & \multicolumn{4}{c}{\texttt{silent\_removal}} & \multicolumn{3}{c@{}}{\texttt{silent\_keep}}\\
Gold & $n$ & Clean & Edited acc. & Answered as & $n$ & Clean & Edited\\
relation & & acc. & [95\% CI] & E1$<$E2/E2$<$E1/Ovl. & & acc. & acc.\\
\midrule
E1$<$E2 & 28 & 1.000 & 0.750 [0.566, 0.873] & 21/0/7 & 26 & 1.000 & 0.960\\
Ovl.    & 23 & 0.913 & 0.609 [0.408, 0.778] & 9/0/14 & 19 & 0.842 & 0.882\\
E2$<$E1 & 30 & 0.900 & 0.433 [0.274, 0.608] & 4/13/13 & 25 & 0.920 & 0.680\\
\midrule
All     & 81 & 0.938 & 0.593 [0.484, 0.693] & 34/13/34 & 70 & 0.928 & 0.836\\
\botrule
\end{tabular}
\footnotetext{E1$<$E2: EVENT1-BEFORE-EVENT2; E2$<$E1: EVENT2-BEFORE-EVENT1; Ovl.: EVENT1-OVERLAP-EVENT2. The \texttt{silent\_removal} editor mentions the diagnosis first, which reverses mention order only for EVENT2-BEFORE-EVENT1 notes. Clean accuracy is on the unedited base notes. Notes answered: \texttt{silent\_removal} clean 80 of 81, edited 81 of 81; \texttt{silent\_keep} clean 69 of 70, edited 67 of 70. Produced by \texttt{per\_class\_silent.py}.}
\end{table}

On \texttt{silent\_keep}, where the note was rewritten by the same editor but the temporal relationship was preserved, BERT's coverage barely moved (0.986 clean to 0.957 silent) and its accuracy fell moderately (0.928 to 0.836 on the same notes, Table~\ref{tab:silent}). The rewrite alone left accuracy unchanged on EVENT1-BEFORE-EVENT2 and overlap notes (0.932 to 0.929) but lowered it on EVENT2-BEFORE-EVENT1 notes (0.920 to 0.680; Table~\ref{tab:silent_perclass}).

On \texttt{silent\_removal}, BERT abstained on 0 of 81 notes: its coverage did not fall (0.988 clean, 1.000 under silent removal) in both the strict and implied-only tiers. Its accuracy on these same 81 notes fell from 0.938 when clean to 0.593 once the temporal evidence was removed, a 34.5-point drop, with the strict tier (no residual cue found by the cue filter or the judge) falling further (0.946 to 0.568) than the implied-only tier (0.930 to 0.614). The gate that was trained to detect \texttt{remove\_evidence} and \texttt{mask\_temporal\_cue} did not generalise to this different form of missing evidence: it kept answering at its normal rate while the answers it gave became substantially less reliable. The silent removal editor always mentions the diagnosis first, which reverses mention order for EVENT2-BEFORE-EVENT1 notes only; silent keep does not reorder. Split by gold relation (Table \ref{tab:silent_perclass}), BERT's accuracy fell in all three classes: EVENT1-BEFORE-EVENT2 from 1.000 to 0.750, overlap from 0.913 to 0.609 and EVENT2-BEFORE-EVENT1 from 0.900 to 0.433. Only 4 of the 30 reordered notes were answered EVENT1-BEFORE-EVENT2, so the reversed order explains little of the drop. Errors followed the cues the edit removed: BEFORE-type notes, stripped of day numbers, were answered as overlap (20 of 58), and overlap notes, stripped of the co-occurrence cue and listed diagnosis-first, were answered EVENT1-BEFORE-EVENT2 (9 of 23). BERT kept answering from the surface cues that remained. Part of the EVENT2-BEFORE-EVENT1 drop reflects sensitivity to rewriting alone, which silent keep shows for that class. On the strict tier, answered accuracy stayed above the 0.354 majority-class rate for both gates (BERT 0.568 [0.409, 0.713]; LLM 0.515 [0.397, 0.632]). Either the judge missed residual evidence the gates use, or the tier's class mix differs from the corpus; the strict tier is not shown to be free of relation-bearing signal, and accuracy there means agreement with the original note's gold relation.

The LLM shows the opposite pattern in the strict tier: coverage fell from 0.756 clean to 0.277 under silent removal (66 of 238 notes answered), a 47.9-point drop, while its \texttt{silent\_keep} coverage was nearly unchanged (0.723 to 0.721). In the implied-only tier, where some order-bearing phrasing survived the edit, its coverage did not fall (0.651 to 0.688) but its accuracy collapsed on the notes it did answer (0.851 to 0.384, barely above the 0.354 majority-class rate), consistent with the gate passing weak or misleading residual cues rather than genuinely abstaining.

\subsection{Provenance graph: abstention-driven discrepancies}
The provenance graph ingested the ungated LLM baseline and the gated LLM outputs for the same 60 held-out notes, giving 120 model-output edges (Table~\ref{tab:kg}). Counts refer to output edges and identical relations from the two configurations appear as two edges. In this pipeline the gated relation comes from the same temperature-0 call as the ungated baseline and the gate only decides whether to release it, so the two can differ only by abstention: all 33 discrepancies were abstentions and the 27 answered notes matched. The experiment shows that the graph stores abstentions as typed, queryable edges with their reasons; it does not test separation of abstentions from relation disagreements, since none could occur. It is not evidence of contradiction detection between independent models and this was not a BERT-versus-LLM comparison.

\begin{table}[t]
\centering\small
\caption{Baseline-versus-gated LLM graph ingestion and disagreement flags.}
\label{tab:kg}
\begin{tabular}{@{}p{7.2cm}p{7.8cm}@{}}
\toprule
Quantity & Value \\
\midrule
Total edges & 120 \\
Abstained edges & 33 (27.5\%) \\
Relation types among non-abstained edges & EVENT1-BEFORE-EVENT2: 36; EVENT1-OVERLAP-EVENT2: 31; EVENT2-BEFORE-EVENT1: 20 \\
Mean coverage (character-span fraction; not comparable to BERT token-mask coverage) & 0.317 \\
Disagreements flagged (baseline versus gated) & 33 (all abstention-driven) \\
\bottomrule
\end{tabular}
\end{table}

\subsection{Secondary analyses: certification and manipulation}
Table~\ref{tab:secondary} gives the perturbation, manipulation and certification results. The BERT certification run used 100 perturbations and $\alpha=0.01$, which requires a lower bound of 0.99 on each of two jointly evaluated bounds at $\alpha/2=0.005$. With 100 of 100 successes the Clopper--Pearson bound is 0.948, below the requirement, so CERTIFIED status was unattainable even for a perfectly stable example. Solving \texttt{beta.ppf(0.005, n, 1)} $\geq 0.99$ shows that 528 perturbations per example are needed. The BERT certification rate of 0.000 therefore mixes this ceiling with severe abstention (28 of 30 examples abstained before perturbation) and measured instability and is not a robustness estimate. The LLM run used $\alpha=0.05$, where the all-success ceiling at $n=100$ is 0.964 and the threshold is reachable (at least 72 perturbations are needed); the LLM baseline certified 5 of 20 examples. The two runs are not comparable.

In the manipulation analysis, 2,000 adversarial variants were generated and 1,887 base-variant pairs remained after restricting to relation-eligible notes and resolvable pairs. These pairs are not restricted to the held-out split: for the BERT models, 1,319 base notes belong to training patients, 279 to development and 289 to test, so the BERT manipulation figures are measured mostly on training notes. The gated BERT model abstained on 99.0\% of them, so its low stability (0.007) mostly reflects abstention. The gated LLM abstained on 43.3\% of adversarial inputs, below its 55.0\% clean abstention rate and left a substantial stability gap among paired evaluations. The natural-variability arm had $n=0$ because the generator produced adversarial variants only.

\begin{table}[t]
\centering\footnotesize
\setlength{\tabcolsep}{4pt}
\caption{Certification and manipulation results for the original gates. Configurations differ in $\alpha$, sample size and evidence representation and are not directly comparable. BERT adversarial pairs include training-patient notes (1,319 of 1,887).}
\label{tab:secondary}
\begin{tabular}{@{}lrrrr@{}}
\toprule
Metric & BERT baseline & BERT gated & LLM baseline & LLM gated \\
\midrule
Certification examples & 30 & 30 & 20 & 20 \\
Certification $\alpha$ & 0.01 & 0.01 & 0.05 & 0.05 \\
Abstained before perturbation & 0/30 & 28/30 & 0/20 & 16/20 \\
Certification rate & 0.000 & 0.000 & 0.250 & 0.000 \\
Mean prediction confidence & 0.926 & 0.042 & 0.939 & 0.158 \\
Mean provenance confidence & 0.567 & 0.033 & 0.850 & 0.158 \\
Adversarial pairs & 1,887 & 1,887 & 60 & 60 \\
Abstain rate (adversarial) & 0.000 & 0.990 & 0.000 & 0.433 \\
Prediction stability & 0.907 & 0.007 & 0.883 & 0.300 \\
Evidence-shift rate & 0.042 & 0.008 & 0.033 & 0.017 \\
\bottomrule
\end{tabular}
\end{table}

\section{Discussion}
The three auditability properties were tested with different strengths of evidence and the differences matter for how the paper should be read.

Compression is the firmest result. Aggregation and revision-representation kept every sampled answer and every mention link at 67\% and 63\% of the node-plus-edge count (77--78\% of serialized size). The retained membership edges are what make this auditable and they are also what cap the saving: aggregation removed nodes only. The fixed-window baseline shows that size alone is the wrong target: it failed on more than half of query points, almost always by returning nothing for an active fact, never with an ABSTAIN. The evidence for the provenance-aware operators is one of preservation, not revision reasoning: every query here is answerable by recency, the operators are lossless by construction, and part of the count-based saving comes from moving mention attributes onto edges. The temporal-approximation operator only partly takes the other course. When an approximated day leaves a query unresolved it returns ABSTAIN (0.4\% of query points), but on a further 1.3\% it returned a wrong answer without a flag. Its rate of unflagged wrong values (1.3\%) is close to the fixed window's (1.2\%); the fixed window's much larger failure is returning nothing (53.4\%). On post-revision points the two reached 0.824 and 0.776. Since the confidence values and revision structure came from the generator, this result does not transfer to noisy records.

The abstention results are more limited than the headline rates suggest. The BERT gate reached 98.7\% clean coverage and 99.0\% abstention recall, a large improvement on the hard-gated model's 96.0\% clean abstention. But every control input contains an explicit marker of omission and pairing each control with its clean note shows that the marker drives the behaviour. One appended sentence moved BERT from answering all 89 notes to answering 3, on notes it classified with 0.843 accuracy when clean. One mask token moved it from answering 96.7\% of 61 notes to answering none. The LLM likewise withheld all 579 notice notes, having answered 58.9\% of them when clean. The gates differ in what they read: BERT reacted fully to either marker, the LLM fully to the sentence and only weakly to the token. Of the two control types, \texttt{remove\_evidence} is the weaker test, being one sentence repeated. In the sense that matters for an audit, this alone does not show that either gate abstains because evidence is missing. The oracle label overstates unavailability in the intact-notice notes and the LLM's 83.5\% overall abstention recall largely measures how often the control text announces the omission.

The marker-free controls of Section~\ref{sec:silentresults} settle the question the marker-carrying controls could not and the two gates answer it differently. BERT showed no sensitivity to silent removal: it abstained on 0 of 81 notes, exactly as on clean input, while its accuracy on those same notes fell 34.5 points (93.8\% to 59.3\%) once the temporal evidence was gone. A light rewrite alone had little effect except on EVENT2-BEFORE-EVENT1 notes (\texttt{silent\_keep}: coverage 98.6\% to 95.7\%, accuracy 92.8\% to 83.6\%). The larger drop under silent removal spans all three relation classes, is not explained by the imposed mention order and follows the temporal cues that were deleted (Table~\ref{tab:silent_perclass}). This is consistent with the gate trained to detect \texttt{remove\_evidence} and \texttt{mask\_temporal\_cue} having learned those markers, or something correlated with them, rather than a transferable notion of evidence sufficiency: without the marker it did not abstain, and it kept answering from whatever surface cues remained. The LLM, by contrast, showed a real, marker-independent response in the strict tier (coverage 75.6\% to 27.7\%, against a near-unchanged 72.3\% to 72.1\% on \texttt{silent\_keep}), though only a partial one and its behaviour in the implied-only tier is harder to interpret: coverage did not fall, but accuracy on the notes it answered collapsed to 38.4\%, barely above the 0.354 majority-class rate, suggesting it passed weak or misleading residual cues rather than genuinely finding sufficient evidence. Neither gate's silent-removal behaviour should be read as solved; the LLM's is a partial and unevenly distributed capability and BERT showed no abstention response at all. The comparison between the two is confounded by supervision (trained versus zero-shot), sample size and evidence representation, so it is not a ranking of model families. The qualitative asymmetry (one architecture generalises somewhat beyond the training markers, the other does not at all) is the main new observation of this evaluation, with one caveat that applies to the LLM only. The strict tier admits only edits from which the same Mistral model, acting as judge, could not determine the relation, so part of the LLM gate's abstention on that tier follows from how the tier was selected. BERT's 0 of 81 result does not depend on this selection.

The recomputed evidence-span recall separates the two gates further, on a different axis. Once scored against gold spans regenerated from the actual paraphrased text, BERT's recall remained low (0.049 ungated, 0.006 gated) and its reconstructed spans were often single sub-word fragments or generic connective phrases the paraphraser itself introduces, not temporal content: this is expected, because its evidence head never received span supervision, and says nothing about BERT's capacity to localize evidence. The LLM's recall on answered notes reached 0.850 with 0.864 precision and its cited spans covered under 80\% of the note in every scorable case, so the LLM did not hedge by quoting most of the note, although notes are short and the reference spans come from the same model family. Read together with the silent-removal result, the LLM cites short spans that match the regenerated evidence when it answers and shows a measurable, if incomplete, ability to notice when that evidence has been removed without warning. For BERT the span result is uninformative because of the training defect, and the silent-removal result shows no abstention response.

Finally, the provenance graph made the abstention visible as data. The 33 flagged discrepancies between the ungated and gated LLM were all abstentions, as that comparison guarantees; the graph stored each with its reason, so an auditor can tell a withheld answer from a missing value. Separating abstentions from genuine disagreements, which needs two independently produced outputs such as two different models, is the diagnostic use the design was built for and has not been tested, nor has ingestion of the control-branch outputs.

\section{Limitations}\label{sec:limitations}
\begin{enumerate}[nosep]
\item The corpus is entirely synthetic. Its clinical realism, lexical diversity, contradiction structure and temporal patterns are bounded by the generator.
\item The paraphrase prompt was not accompanied by an independent fact-preservation audit.
\item Relation labels, evidence spans and availability labels are generator-produced. No manually reviewed clinical annotation exists.
\item The evidence-span recall metric of Table~\ref{tab:orig} is invalid as implemented: gold spans are searched by exact substring in paraphrased notes and the value defaults to 0.0. This is an evaluation-harness defect, not a demonstrated model failure. Section~\ref{sec:recall2} reports a corrected metric.
\item Every evidence-unavailable control carries an explicit marker of omission (a fixed sentence, an appended notice or a mask token). \texttt{remove\_evidence} is one distinct input repeated, so its 100\% abstention is one observation. Pairing controls with their clean notes shows both gates responding to the marker. The results show sensitivity to signalled absence and cannot be read as detection of silent absence.
\item In 579 of 1,000 \texttt{mask\_temporal\_cue} notes the original text is retained with an appended notice and in the other 421 a mask token replaces one cue phrase while other temporal wording may remain. The oracle label \texttt{evidence\_available}$=0$ describes the intervention and not verified absence; BERT's clean accuracy of 0.843 on the intact-notice notes indicates the relation was recoverable there.
\item The availability-supervised BERT gate was trained on controls produced by the same generator functions that produce the test controls, so its control results are in-distribution by construction. Its evidence head was trained on all-zero span labels and no development-set model selection was used.
\item The unsafe-answer rate counts answered inputs whose relation differs from the gold relation of the original note. On \texttt{mask\_temporal\_cue} it can therefore understate unsafe answering (a chance-correct answer is not counted) and, where evidence remains, overstate it. Coverage on unavailable evidence is reported alongside.
\item An earlier LLM run without EVENT1/EVENT2 definitions in the prompt answered clean notes at close to the majority-class rate (Table~\ref{tab:clean}). All other LLM results use the prompt with definitions; the earlier original-gate LLM results (Tables~\ref{tab:orig}, \ref{tab:kg} and~\ref{tab:secondary}) were not rerun with it.
\item The silent-removal and silent-keep controls (Section~\ref{sec:silent}) and the regenerated gold spans (Section~\ref{sec:goldspans}) were both produced with the same model family that is evaluated as the LLM gate, using it as editor, relocator and judge. A second, independently sourced judge or editor model was not available. A human audit sample was written by the generation scripts but has not been completed at the time of writing. In particular, the strict silent-removal tier is defined by the judge failing to infer the relation, so the LLM gate's lower coverage on that tier is partly a consequence of tier selection rather than independent evidence of sensitivity. Silent controls were also pre-filtered to notes whose original relation the judge recovered.
\item Notes for which \texttt{silent\_removal} or the gold-span regeneration succeeded (507 of the 561 judge-validated notes and 482 of 1,000 respectively) are not a random subset: both procedures require enough narrative content for an editor or relocator to work with, so the accepted subset likely skews toward longer, more detailed notes. The recall and silent-removal results in Sections~\ref{sec:recall2} and~\ref{sec:silentresults} should be read as applying to that subset, not as corpus-wide estimates; this was not directly checked (for example by comparing note length between accepted and rejected notes).
\item BERT's reconstructed evidence spans (Section~\ref{sec:recall2}) were qualitatively inspected only for the ungated model; whether the availability-supervised (gated) model's much lower recall reflects the same pattern of sub-word and generic-connective spans, or a different failure mode, was not examined. No LLM-style precision or cited-span-length metric was computed for BERT, so its recall cannot be checked for the boilerplate-citation effect that the LLM's precision and span-length figures rule out.
\item In the \texttt{silent\_removal} implied-only tier, the LLM's coverage did not fall relative to clean, but its answered accuracy fell to 38.4\%, barely above the 0.354 majority-class rate; this pattern (passing weak or misleading residual cues) is inferred from the aggregate numbers and was not traced to individual cases.
\item The BERT and LLM evaluations differ in supervision, sample size (150 versus 1,000 clean notes; 300 versus 2,000 controls) and evidence representation; BERT token coverage and LLM character coverage are not comparable. The earlier 60-note LLM results have wide intervals (accuracy 0.617, Wilson 95\% interval 0.490--0.729).
\item The original hard-gated BERT encoder was frozen; the effect of fine-tuning was not evaluated.
\item Certification runs used different $\alpha$ values and sample counts. The BERT threshold was unattainable at 100 perturbations.
\item Compression confidence values were simulated from note-style heuristics, not calibrated extraction confidence.
\item The natural-variability arm had no evaluated examples and the position-control diagnostic was unavailable because the generator produced no reverse-order admission variants.
\item The provenance-graph experiment compared LLM baseline and LLM gated outputs; cross-model contradiction detection and ingestion of control-branch outputs were not evaluated. The contradiction control variants were not run.
\item The BERT manipulation analysis (Table \ref{tab:secondary}) was run on adversarial pairs from all patients, most of them in the training partition; it was not repeated on the 289 test-patient pairs.
\item No clinician evaluation, workflow evaluation or prospective validation was conducted. Intervals are reported only for the new abstention metrics and treat inputs as independent; other results are point estimates from small samples and need replication.
\item The clean relation task has a template shortcut: a no-learning rule using mention order and day-number count reaches 0.651 accuracy (Section~\ref{sec:corpus}), so clean accuracies overstate temporal reasoning.
\item \texttt{silent\_keep} is not matched for reordering and restructuring, and it alone lowered BERT's accuracy on EVENT2-BEFORE-EVENT1 notes (0.920 to 0.680).
\item Every compression query is answerable by recency, so the experiment tests information preservation, not revision reasoning; size is a node-plus-edge count, and serialized size shows a smaller saving.
\item The v2 LLM re-query check tests consistency under rewording, not necessity.
\item BERT input is truncated at 128 tokens, so an appended notice can fall outside the input for the longest notes.
\end{enumerate}

\section{Conclusion and Future Work}\label{sec:conclusion}
 
On synthetic data, provenance-aware compression kept every sampled answer and mention link at 63--67\% of the node-plus-edge count (77--78\% of serialized size), and the provenance graph stored abstentions as queryable edges with their reasons. This part of the pipeline is the firmest result in the study, with two qualifications: every query was answerable by recency, so the experiment tests preservation of the information needed to answer rather than revision reasoning, and the operators are lossless by construction.
 
Evidence-availability abstention is more mixed. On controls carrying an explicit marker, both gates responded mainly to the marker: one appended sentence or one mask token was enough to change BERT's output almost entirely, and the LLM responded fully to the sentence but only weakly to the token. Marker-free controls, paired with a minimal rewrite that preserved the evidence, separated the two gates further. BERT abstained on none of 81 notes whose temporal evidence had been silently removed, while its accuracy on those notes fell 34.5 points. The drop spanned all three relation classes and followed the deleted cues, which is consistent with a gate that learned the training markers rather than evidence sufficiency. The LLM showed a partial and unevenly distributed capacity to notice that evidence was missing without being told so, with coverage falling from 75.6\% to 27.7\% in the strict tier, a tier the same model selected as judge. Recomputed against 482 regenerated gold spans, the LLM's character-span recall on answered notes reached 0.850 with 0.864 precision, with cited spans covering about a third of the note; BERT's recall was near zero, reflecting an evidence head trained without span labels.
 
Several priorities from an earlier version of this work have been addressed: marker-free controls with a matched negative control were built and run on both gates (Section~\ref{sec:silentresults}), and gold spans were regenerated well enough to report a corrected, non-degenerate recall metric. What remains open follows from the limitations of that new work (Section~\ref{sec:limitations}). A second, independently sourced model should serve as editor and judge for the silent-removal and gold-span pipelines, and the human audit sample should be completed. The marker-free negative control should be rebuilt with the same reordering and restructuring as the removal edit, which the present one does not apply. BERT's evidence head should be retrained on the regenerated spans, its gated spans inspected as the ungated model's were, and a precision or span-length metric added so its recall can be checked for boilerplate citation. Whether the accepted, longer-narrative subset of notes represents the corpus should be checked directly. The relation task should be rebuilt without the template shortcut, and compression should be tested on queries that require traversing revision links, such as out-of-order documentation. Certification should be repeated with matched parameters and at least 528 perturbations at $\alpha = 0.01$, and simulated confidence replaced by model-derived confidence. Finally, the corrected pipeline should be evaluated on real longitudinal records with expert temporal and evidence-span annotation.
 
Until then the system is a research prototype for provenance and graph-mechanism evaluation. The abstention mechanism in particular should be read as a partly supported hypothesis rather than a solved problem: the evidence is consistent with gates trained or prompted on marker-carrying controls learning to detect the marker and, in the LLM's case only, showing some transfer to genuinely missing evidence.

\section*{Declarations}
\textbf{Funding:} This work did not receive any external funding.\\
\textbf{Ethics approval:} This study used only programmatically generated synthetic clinical text; no real patient data were used and no human subjects were involved.\\
\textbf{Data availability:} The synthetic corpus is available at \url{https://github.com/JudithJeyafreeda/Provenance-Aware-TRE/}. The release includes the generated notes, facts, revision edges and adversarial variants. Model outputs, control files, regenerated gold spans and evaluation results are not distributed; they are produced by the released scripts. The marker-carrying evidence-unavailable control file is regenerated from the notes with \texttt{make\_evidence\_unavailable\_adversarial.py}; the marker-free \texttt{silent\_removal}/\texttt{silent\_keep} controls and the regenerated gold-span file are produced by \texttt{make\_silent\_removal\_controls.py} and \texttt{regenerate\_gold\_spans.py} respectively and are not yet archived as static releases.\\
\textbf{Code availability:} Code, execution instructions, dependency information, model configurations and data-generation materials are in the same repository, including the availability-supervised BERT scripts, the marker-free control and gold-span-regeneration scripts and the recall and stratification scripts listed in Table~\ref{tab:scripts}.\\
\textbf{Author contributions:} J.J.A. conceived the study, developed the software, generated the synthetic corpus, performed the analyses, interpreted the results and wrote the manuscript.\\
\textbf{Competing interests:} The author declares no financial or non-financial competing interests.
\bibliography{sn-bibliography}

\appendix
\section{Reproducibility and implementation details}
The complete source code is in \url{https://github.com/JudithJeyafreeda/Provenance-Aware-TRE/}. The repository should be archived as a versioned release with a persistent identifier matching this manuscript. The corpus contains 1,000 patients, 3,353 notes, 2,220 clinical facts, 220 revision edges and 2,000 adversarial variants. The original hard-gated BioClinicalBERT encoder was frozen. LLM extraction used \texttt{mistralai/Mistral-Small-3.2-24B-Instruct-2506} at temperature 0.0 through an OpenAI-compatible endpoint. The compression evaluation used a confidence threshold of 0.65, a fixed-window size of $k=2$ and six sampled query days per patient and concept type. Table~\ref{tab:scripts} maps scripts to roles; file names follow the repository.

\begin{table}[h]
\centering\footnotesize
\caption{Script-to-role mapping.}
\label{tab:scripts}
\begin{tabular}{@{}p{6.6cm}p{8.8cm}@{}}
\toprule
Script & Role \\
\midrule
\texttt{sythetic\_data\_generation.py} & Corpus generation (Section 2.2) \\
\texttt{provenance\_first\_Bio\_Clinical\_BERT.py} & Hard-gated BERT extraction, certification (Sections 2.6, 2.10) \\
\texttt{provenance\_first\_LLM.py} & Gated LLM extraction, certification (Sections 2.6, 2.10) \\
\texttt{compression\_operators.py} & Evidence Graph compression operators and evaluation, including serialized size and revision subsets (Section 2.7) \\
\texttt{provenance\_temporal\_kg.py} & Provenance graph, entity resolution, contradiction flags, federation-ready aggregation (Section 2.9) \\
\texttt{make\_evidence\_unavailable\_adversarial.py} & Evidence-unavailable (marker-carrying) control generation (Section 2.3) \\
\texttt{make\_silent\_removal\_controls.py} & Marker-free \texttt{silent\_removal}/\texttt{silent\_keep} control generation: LLM editor, judge validation, strict/implied tiers (Section 2.4) \\
\texttt{regenerate\_gold\_spans.py} & Post-paraphrase gold-span regeneration: LLM relocator, judge validation, verified/relation-only tiers (Section 2.5) \\
\texttt{bert\_control\_inference.py} & Coverage-gated BERT training and control inference \\
\texttt{train\_availability\_supervised\_bert.py} / \texttt{\_v2.py} / \texttt{\_v3.py} & Availability-supervised BERT training; v2 adds save/load of trained weights and scoring of new control files without retraining; v3 adds reconstruction of per-row predicted evidence spans and token-overlap recall against regenerated gold spans (Section 2.6) \\
\texttt{llm\_availability\_gated\_inference\_v1.py} / \texttt{\_v2.py} & LLM gate with availability-aware prompt; v2 adds EVENT1/EVENT2 definitions to the prompt and exports the cited evidence span per row \\
\texttt{compute\_llm\_evidence\_recall.py} & LLM character-span recall, precision and cited-span-length against regenerated gold spans (Section 2.5, results in Section~\ref{sec:recall2}) \\
\texttt{evaluate\_control\_abstain\_v1.py} & Abstention metrics on clean and control predictions (Section 2.8) \\
\texttt{stratify\_masked\_cue.py} & Split of control results by variant and (for \texttt{mask\_temporal\_cue}) by \texttt{masked\_cue}, with optional paired clean coverage; extended to score any control variant, including \texttt{silent\_removal}/\texttt{silent\_keep} (Section 2.8) \\
\texttt{per\_class\_silent.py} & Split of \texttt{silent\_removal}/\texttt{silent\_keep} results by gold relation, paired with clean predictions (Section~\ref{sec:silentresults}) \\
\texttt{README.md} & Installation, execution instructions, configuration \\
\bottomrule
\end{tabular}
\end{table}
\end{document}